# CrossStateECG: Multi-Scale Deep Convolutional Network with Attention for Rest-Exercise ECG Biometrics

Dan Zheng[1][0009-0008-3588-9715], Jing Feng[2,1(✉)][0000-0002-3915-4926] and Juan Liu[2,1][0000-0001-9344-7415]

[1] Key Laboratory of Aerospace Information Security and Trusted Computing, Ministry of Education , School of Cyber Science and Engineering ,Wuhan University
[2] School of Artificial Intelligence, Wuhan University, Wuhan, China
gfeng@whu.edu.cn

**Abstract.** Current research in Electrocardiogram (ECG) biometrics mainly emphasizes resting-state conditions, leaving the performance decline in rest-exercise scenarios largely unresolved. This paper introduces CrossStateECG, a robust ECG-based authentication model explicitly tailored for cross-state (rest-exercise) conditions. The proposed model creatively combines multi-scale deep convolutional feature extraction with attention mechanisms to ensure strong identification across different physiological states. Experimental results on the exercise-ECGID dataset validate the effectiveness of CrossStateECG, achieving an identification accuracy of 92.50% in the Rest-to-Exercise scenario (training on resting ECG and testing on post-exercise ECG) and 94.72% in the Exercise-to-Rest scenario (training on post-exercise ECG and testing on resting ECG). Furthermore, CrossStateECG demonstrates exceptional performance across both state combinations, reaching an accuracy of 99.94% in Rest-to-Rest scenarios and 97.85% in Mixed-to-Mixed scenarios. Additional validations on the ECG-ID and MIT-BIH datasets further confirmed the generalization abilities of CrossStateECG, underscoring its potential as a practical solution for post-exercise ECG-based authentication in dynamic real-world settings.

## 1 Introduction

With the advancement of biometric recognition technologies, identity authentication has become increasingly crucial in modern information security systems. Although traditional biometric methods such as fingerprints, facial recognition, and iris scanning are widely used, they are vulnerable to theft and forgery. ECG, as a physiological signal, is emerging as a significant research area in biometric recognition due to its uniqueness and the difficulty of being forged.

In 2001, Biel et al. [1] first proposed using ECG for human identification, demonstrating individual uniqueness by analyzing 30 temporal-domain features extracted from the 12-lead ECG signals of 20 subjects, and emphasized that single-lead features were sufficient for identification. Subsequently, in 2002, Shen et al. [2] further validated the feasibility of individual identification using single-lead ECG signals through template matching and decision-based neural networks, establishing a foundation for ECG-based biometric authentication. In 2005, Israel et al. [3] introduced multiple temporal-domain features, successfully achieving individual identification and heartbeat classification using linear discriminant analysis. This further confirmed the viability of ECG signals as biometric traits from various perspectives.

ECG identification methodologies have evolved from fiducial-point-based approaches to non-fiducial-point methods as research advanced. In 2016, Hejazi et al. [4] proposed a kernel-based non-fiducial-point ECG biometric authentication method that tackled limitations inherent in traditional fiducial-point detection techniques, paving the way for subsequent non-fiducial methods. Regarding machine learning applications, Liu et al. [5] developed a generalized regression neural network (GRNN)-based method in 2018, which outperformed conventional KNN and SVM approaches. In 2019, Liu et al. [6] further investigated the use of Long Short-Term Memory (LSTM) networks, significantly improving the accuracy of ECG-based identification. In 2020, Hsu et al. [7] innovatively transformed ECG signals into two-dimensional images and utilized transfer learning techniques based on pre-trained AlexNet and ResNet18 models, achieving remarkable performance in ECG-based identification.

The rapid development of deep learning technology has led to breakthroughs in ECG-based identity recognition. In 2021, Liu et al. [8] proposed a two-stage feature fusion model. The first stage combined Hilbert transform and power spectrum features, while the second stage utilized PCANet for deep feature extraction. A MaxFusion algorithm was introduced to merge features from different PCANet layers, achieving an identification accuracy of 99.77% on a combined ECG-ID, MIT-BIH, and PTB dataset. In 2022, Yang et al. [9] designed a Feature Reuse Residual Network (FRRNet), which integrates max pooling and average pooling within residual networks to better preserve useful features and reduce noise interference. They combined outputs from both deep and shallow network layers to tackle the accuracy reduction caused by noise. In 2023, Zhang et al. [10] introduced a hybrid feature extraction and sparse representation method that combines fixed ECG and frequency-specific features, thus improving recognition accuracy.

It is important to note that the studies mentioned above primarily validated their approaches using publicly available databases, such as MIT-BIH and ECG-ID, which only contain ECG signals collected during resting states. While significant advancements have been made in resting-state ECG-based identity recognition, research on identifying individuals from ECG signals recorded during physical activity remains limited. In post-exercise scenarios, ECG signals undergo substantial fluctuations due to physiological changes. Factors like heart rate variability can modify the morphological characteristics of ECG signals, posing considerable challenges for identification systems. Therefore, developing an ECG-based identification method that adapts to varying physiological conditions is of significant theoretical and practical importance.

In response to the characteristics of resting and post-exercise ECG signals, this study makes the following contributions:

1. A cross-state ECG authentication model focused on modified multi-similarity metric learning and deep feature convolution. This design aims to investigate the effect of post-exercise states on ECG signal features, an area that has been less explored in prior research. The model was developed and validated across several experimental scenarios, with results indicating that the authentication accuracy remains consistently high.
2. An adaptive authentication strategy designed for practical applications. Innovatively combining global statistical features with individual differences created a multi-level threshold mechanism to dynamically adjust the authentication threshold during the transition from resting to exercise states.

## 2      Related Work

In the research on ECG-based biometric authentication in post-exercise scenarios, Sung et al. [17] proposed a feature fusion method utilizing Linear Discriminant Analysis (LDA) to systematically validate the recovery characteristics of ECG signals following high-intensity exercise and the feasibility of identity authentication. Deep metric learning has seen significant advancements in biometric recognition in recent years. Wang et al. [12] introduced the Multi-Similarity Loss (MS Loss) in image recognition for the first time. This method creatively combines self-similarity, positive-relative similarity, and negative-relative similarity of samples, leading to substantial performance improvements in tasks such as image retrieval. Inspired by this, Zhao et al. [11] applied MS Loss to ECG-based identity recognition. They performed exploratory research on the mutual authentication of ECG signals in two physiological states: post-exercise and rest. This study offers a fresh perspective on addressing the impact of physiological state changes on ECG recognition performance. Zhao et al.' experimental results reveal a significant dip in recognition performance in cross-validation scenarios before and after exercise, underscoring the complexity and challenges of ECG-based identity recognition under varying physiological states. Saleh et al. [21] significantly improved the robustness and accuracy of electrocardiogram recognition in different physiological states, especially after exercise, by combining a dual expert architecture and domain adaptation. The recognition rate was 68.95% in the initial recovery phase and 86.45% in the late recovery phase. There has been relatively little research in the field of electrocardiogram recognition in recent years, especially in terms of performance improvement in physiological states after exercise. Therefore, this article mainly cites earlier classic studies in the literature review and proposes innovative recognition methods based on current technological developments, aiming to solve the robustness problem of electrocardiogram recognition in post exercise states and fill the research gap.

## 3    Method

### 3.1    Dataset and Preprocessing

**Dataset Description.** This study employs the exercise-ECGID dataset [14] as the primary experimental data. Collected and publicly released by the laboratory at South China University of Technology, this dataset holds unique value as it provides paired ECG recordings of the same subject in both resting and post-exercise states. The dataset includes ECG recordings from 45 healthy volunteers (33 males and 12 females, aged 18–22 years), collected using a wearable wrist-based lead II configuration with a sampling frequency of 300 Hz. Data collection for each subject consists of two phases: (1) a baseline ECG recording of approximately 5 minutes in a resting state, with an average heart rate of about 70 beats per minute (bpm); (2) a post-exercise ECG recording of around 150 seconds, obtained immediately after completing a standardized exercise protocol that includes steady-state running and stair climbing, with a heart rate range of 90–150 bpm. This paired data collection approach ensures consistency and reliability in cross-state research.

To further evaluate the generalization capability of the proposed method, we utilize two widely recognized standard ECG datasets: the MIT-BIH Normal Sinus Rhythm Database [13] and the ECG-ID Database [18]. This multi-dataset validation approach supports a comprehensive evaluation of the model's generalization ability and aids in assessing its adaptability across various data collection environments and physiological states.

**Signal Preprocessing.** This study develops a multi-stage signal processing pipeline to ensure ECG signal quality and consistency. First, a 4th-order Butterworth bandpass filter (0.5–40 Hz) is applied for signal denoising. The choice of this frequency band relies on the primary spectral characteristics of ECG signals, effectively preserving the morphological integrity of the QRS complex while suppressing noise and motion artifacts. A 1st-order Butterworth high-pass filter (cutoff frequency: 0.5 Hz) is used for baseline drift correction to enhance baseline stability further. Subsequently, Z-score normalization is applied to standardize signal amplitudes across datasets, ensuring comparability of signals across different acquisition environments.

An enhanced Hamilton-Tompkins algorithm is applied for signal segmentation and QRS complex detection. Given the substantial increase in heart rate due to exercise (90–150 bpm compared to 70 bpm at rest), we propose an adaptive segmentation strategy in which resting-state ECG signals are divided into 6-second segments (1,800 samples) centered on R-peaks. Simultaneously, post-exercise ECG signals are segmented into 4-second segments (1,200 samples) centered on R-peaks. The R-peak is consistently located at 25% of the segment length, which improves the capture of comprehensive electrocardiographic dynamics. To ensure the reliability of the extracted signals, a multi-level quality control framework is established.

This study employs a rigorous subject-independent partitioning strategy for data organization to prevent identity leakage. Specifically, in the Exercise-ECGID dataset, resting-state data are split into training (80%) and validation (20%) sets. In contrast, post-exercise data serves as an independent test to assess cross-state generalization. The same partitioning strategy is applied to the MIT-BIH and ECG-ID datasets. A focused data augmentation strategy is introduced to tackle class imbalance in the ECG-ID dataset. Furthermore, all datasets undergo a standardized preprocessing pipeline to ensure reliable cross-dataset validation, which includes filtering parameters, segmentation strategies, and quality control criteria. This harmonized data preprocessing approach establishes a fair foundation for subsequent feature learning and model evaluation.

### 3.2 Model

This study proposes an end-to-end deep learning model to address the challenge of cross-physiological state ECG identity authentication. The core of this model lies in the robust representation of ECG signals across different physiological states through the organic integration of multi-scale feature extraction, deep feature learning, and attention mechanisms. As shown in Figure 1, the network takes single-lead ECG signals as input and employs a multi-branch parallel structure. Each branch independently captures time-domain features at different scales, and through feature fusion and attention enhancement mechanisms, it extracts discriminative identity features.

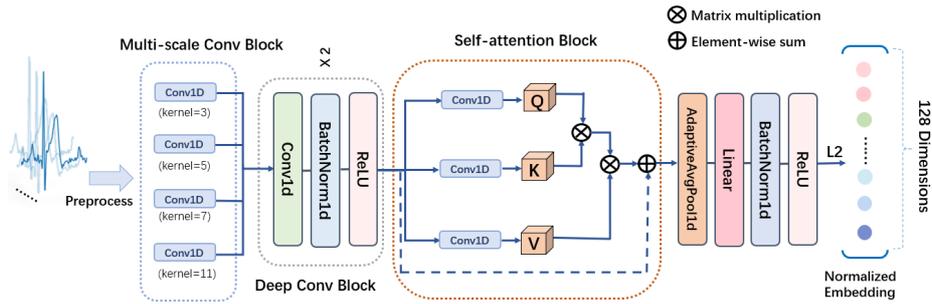

**Fig. 1.** Framework of our proposed CrossStateECG biometric model

In the multi-scale feature extraction module, this study designs four parallel one-dimensional convolution branches with kernel sizes of 3, 5, 7, and 11, each producing 64 feature channels. This progressive receptive field design has clear physiological justification: the small kernel (k=3) focuses on capturing local morphological features such as the QRS complex; the medium-sized kernels (k=5, 7) are responsible for extracting intermediate-scale waveform features like the P-wave and T-wave; the large kernel (k=11) helps capture long-range dependencies between heartbeats. By employing this multi-scale feature extraction strategy, the network can simultaneously model the dynamic characteristics of ECG signals at different time scales, significantly enhancing its adaptability to signal variations during exercise states.

To further enhance the capability of feature representation, the network introduces a deep feature learning module consisting of two consecutive convolution blocks. Each convolution block includes a one-dimensional convolution layer (kernel_size=3), a batch normalization layer, and a ReLU activation function. The first convolution block maintains the input dimension [B, 256, L], while the second block increases the feature channels to 512, improving the network's feature extraction capacity. This deep structural design allows the network to progressively abstract and extract more discriminative identity features from the raw ECG signals.

Considering the susceptibility of ECG signals to noise interference following exercise, this study introduces a self-attention enhancement module after feature extraction. This module first projects the input features into the Query (Q) and Key (K) spaces using a 1×1 convolution, reducing the dimensionality to 1/8 of the original size (i.e., 64 channels). In contrast, the Value (V) features maintain 512 channels to preserve sufficient feature information. After calculating attention weights through matrix multiplication, a learnable scaling factor (γ) and residual connection are introduced, allowing the network to adaptively adjust the influence of the attention mechanism, emphasizing key identity features while retaining the original features.

In the final stage of the network, a "pooling-mapping-normalization" feature processing pipeline is used. First, global average pooling compresses the temporal dimension, followed by a fully connected layer that maps the 512-dimensional features into a 128-dimensional compact embedding space. The introduction of L2 normalization ensures a consistent scaling of the features, which is particularly crucial for subsequent similarity measurements. The final identity classifier performs identity prediction based on the normalized features.

### 3.3   Loss Function Design

To improve the performance of ECG-based identity authentication, we propose a multi-task loss function that combines Focal Loss[20] $L_{focal}$ and improved Multi-Similarity Loss $L_{ms}$. The total loss function is defined as:

$$L_{total} = \alpha L_{focal} + (1-\alpha) L_{ms} \quad (1)$$

This study assigns a larger weight to the modified Multi-Similarity Loss, as feature metric learning plays a more critical role in identity authentication tasks. In contrast, the Focal Loss is given a smaller weight and serves as auxiliary supervision to alleviate the class imbalance issue in ECG data after exercise. Experimental results show that when $\alpha = 0.05$, the model achieves optimal performance on the validation set.

### 3.4   Based on Improved Multi-Similarity Metric Learning

This paper draws on the concept of Multi-Similarity Loss proposed by Wang et al. [12] and adapts it for the ECG identity authentication task. This study introduces a numerical stability parameter $\epsilon$ to address the variability of ECG signals under different physiological states and employs a truncation operation to limit the range of similarity differences. This approach effectively prevents gradient explosion and vanishing gradient issues during training. The improved Multi-Similarity Loss is defined as follows:

$$L_{ms} = \frac{1}{N}\sum_{i=1}^{N}\left(L_{pos}^{i}+L_{neg}^{i}\right) \quad (2)$$

The loss functions for positive sample pairs and negative sample pairs are defined as follows:

$$L_{pos}^i = \frac{1}{\beta_p} \log\left(1 + \sum_{j \in P_i} e^{-\beta_p(d_{ij})}\right) \tag{3}$$

$$L_{neg}^i = \frac{1}{\beta_n} \log\left(1 + \sum_{k \in N_i} e^{\beta_n(d_{ik})}\right) \tag{4}$$

$d_{ij}$ and $d_{ik}$ represent the cosine similarity values after being truncated by hyper-parameters $\tau$:

$$d_{ij} = \max(-\tau, \min(\tau, s_{ij} - \lambda)) \tag{5}$$

A positive sample pair $(i, j)$ is defined as a sample pair with the same identity, while a negative sample pair $(i, k)$ is defined as a sample pair with different identities. $P_i$ represent the positive sample set with the same identity as sample $i$ and $N_i$ represent the negative sample set with a different identity. $\lambda$ is used to distinguish the similarity threshold between positive and negative sample pairs, with a default value of 0.5. The hyperparameters $\beta_p$ and $\beta_n$ are scaling factors for the positive and negative sample pairs, controlling the weights of the positive and negative samples.

$s_{ij}$ represents the cosine similarity between the normalized feature vectors $\mathbf{f_i}$ and $\mathbf{f_j}$ of the positive sample pair $(i, j)$, computed with numerical stability as follows

$$s_{ij} = \frac{f_i^T f_j}{|f_i||f_j| + \epsilon} \tag{6}$$

### 3.5 Adaptive Authentication Strategy

In response to the variability of ECG signals under different physiological states, this paper proposes a multi-level adaptive threshold system that dynamically adjusts authentication thresholds based on the user's feature distribution. The system incorporates global, personal, and local reference mechanisms to determine the final authentication threshold by comprehensively considering feature distributions at different scales. The adaptive threshold system first establishes three levels of reference metrics.

**Global Statistical Features.** Compute the statistical characteristics of genuine and impostor scores across the entire training set.

$$F_g = \frac{\tau_b - \mu_i}{\mu_g - \mu_i} \tag{7}$$

$\tau_b$ represents the baseline threshold, while $\mu_g$ and $\mu_i$ denote the global mean scores of genuine and impostor samples respectively.

**Personal Feature Statistics.** Personalized features are extracted for each user p. $\sigma_g$ is the global standard deviation of the genuine scores.

$$F_p = \frac{\mu_p - \mu_g}{\sigma_g} \qquad (8)$$

**Local Distribution Features.** Local distribution information is captured using percentiles. $S = \{s_1, s_2, ..., s_N\}$ represent the set of all similarity scores (including both genuine and impostor scores). The local distribution features are defined using the quantile function Q(p) as follows.

$$Q(p) = \inf\{x \in S : F(x) \geq p\} \qquad (9)$$

the key quantile points are selected as $p \in \{0.25, 0.50, 0.75\}$. The interpolation function is expressed as:

$$F_l = \text{interp}(\tau_b, Q, [0.2, 0.4, 0.6]) \qquad (10)$$

The final adaptive threshold $\tau_p$ is calculated using a three-factor weighting scheme

$$\tau_p = \tau_b \cdot \left(w_g(1+F_g) + w_p(1+F_p) + w_l F_l\right) \qquad (11)$$

$w_g$, $w_p$, and $w_l$ represent the weight coefficients for the global, personal, and local factors, respectively.

The adaptive strategy proposed in this paper demonstrates unique advantages, especially in handling cross-state ECG identity recognition problems. This strategy, through a global-personal-local three-layer architecture, not only maintains the overall stability of the system but also provides necessary personalized adjustment capabilities. Moreover, the strategy has low computational overhead and is simple to implement, making it highly suitable for deployment in practical applications. These characteristics allow the strategy to show significant practical value in addressing the challenge of cross-state ECG identity recognition.

## 4   Experiments

### 4.1   Experimental Setup

A series of systematic experiments was designed to comprehensively evaluate the performance of the proposed method in cross-physiological state ECG identity authentication. These experiments are based on three publicly available datasets and the exercise-ECGID dataset, encompassing various authentication scenarios during rest and post-exercise states. Since the three publicly available datasets do not include ECG signals from the post-exercise state, the exercise-related experiments are conducted solely on the exercise-ECGID dataset.

All experiments were conducted on a workstation equipped with an NVIDIA RTX 3090 GPU. The model was trained using the Adam optimizer with an initial learning rate of 0.0004, and the ReduceLROnPlateau strategy was used for learning rate adjustment. The batch size was set to 32, resulting in a total of 200 training epochs. To ensure reproducibility, all random seeds were fixed at 42. The dataset for each experiment was divided into training, validation, and test sets.

## 4.2 Experimental Results and Analysis Indicators

To systematically evaluate the performance of the proposed method in cross-physiological state ECG identity authentication, three groups of seven experimental scenarios were designed, covering same-state recognition, cross-state recognition, and cross-dataset validation.

In ECG-based identity authentication systems, other measurement indicators are also introduced in addition to accuracy (ACC). False Acceptance Rate (FAR) measures the probability that the system incorrectly classifies an unauthorized user as legitimate; a high FAR may pose security risks. False Rejection Rate (FRR) represents the probability that the system erroneously rejects a legitimate user, which, if too high, may negatively impact user experience. Area Under the Curve - Receiver Operating Characteristic (AUC-ROC) is used to evaluate the discriminative capability of the classifier, where an AUC value closer to 1 indicates a better ability to distinguish between legitimate and unauthorized users, while a lower AUC suggests poor recognition performance. These metrics collectively assess the balance among security, reliability, and user experience in the system.

**Same-State Recognition Experiments.** This set of experiments seeks to evaluate the model's performance in a specific physiological state.
*Resting State Recognition (Rest2Rest).* The model is trained and tested using ECG data from the resting state to validate its recognition ability under ideal conditions. The training and test sets are composed of ECG signals collected in the resting state.
*Post-Exercise State Recognition (Exercise2Exercise).* The model is trained and tested using ECG data from the post-exercise, evaluating its ability to handle ECG signals in the post-exercise. The training and test sets are composed of ECG signals collected after exercise.
*Mixed State Training (Mix2Mix).* This scenario uses 50% of ECG data from the resting state and 50% from the post-exercise for training, and testing is performed in both physiological states.

Table 1. Performance Evaluation of Same-State ECG Identity Authentication

| Methods | Datasets | Train-test mode | Number of subjects | Acc (%) | FAR (%) | FRR (%) | AUC-ROC(%) |
|---|---|---|---|---|---|---|---|
| Proposed | exercise-ECGID | Rest2Rest | 45 | 99.94 | 0.00 | 0.12 | 100 |
| Proposed | exercise-ECGID | Exercise2Exercise | 45 | 99.86 | 0.00 | 0.28 | 100 |
| Proposed | exercise-ECGID | Mix2Mix | 45 | 97.85 | 0.27 | 4.56 | 98.85 |

**Cross-State Recognition Experiments.** This set of experiments evaluates the generalization ability of the model in physiological state transition scenarios.

*Rest to Post-Exercise (Rest2Exercise).* The model is trained on ECG data from a rest state and tested on data from a post-exercise state. This experiment simulates the most common scenario in practical applications, where registration is done in the resting state and recognition occurs in the post-exercise state.

*Post-Exercise to Rest (Exercise2Rest).* The model is trained on ECG data from a post-exercise state and tested on data from a rest state. This experiment validates the bidirectional adaptability of the model's feature extraction.

**Table 2.** Performance Evaluation of Cross-State ECG Identity Authentication

| Methods | Datasets | Train-test mode | Number of subjects | Acc (%) | FAR (%) | FRR (%) | AUC-ROC(%) |
|---|---|---|---|---|---|---|---|
| Saleh et al. [21] | UofTDB | Rest2Exercise | 47 | 86.45 | / | / | / |
| Wang et al. [14] | exercise-ECGID | Rest2Exercise | 45 | 61.40 | / | / | / |
| **Proposed** | exercise-ECGID | Rest2Exercise | 45 | **92.50** | 1.61 | 13.33 | 98.72 |
| Proposed | exercise-ECGID | Exercise2Rest | 45 | 94.72 | 8.95 | 1.61 | 99.46 |

**Cross-Dataset Recognition Experiments.** This set of experiments verifies the generalizability and scalability of the method and compares its performance with existing methods.

*ECG-ID Dataset Validation.* Training and testing are conducted on the ECG-ID dataset to evaluate the method's performance in long-term ECG variations. This dataset includes ECG records from different periods, allowing the assessment of the model's robustness to temporal changes.

*MIT-BIH Dataset Validation.* Training and testing are conducted on the MIT-BIH dataset to assess the method's adaptability to different types of ECG signals. This dataset features a variety of rhythm change patterns, enabling the evaluation of the model's ability to handle ECG dynamics changes.

**Table 3.** Performance Comparison With The State-of-the-art Works

| Methods | Datasets | Train-test mode | Number of subjects | Acc (%) |
|---|---|---|---|---|
| Saleh et al. [21] | UofTDB | Rest2Rest | 47 | 98.12 |
| Wang et al. [14] | exercise-ECGID | Rest2Rest | 45 | 93.80 |
| Kim et al. [15] | MIT-BIH | Rest2Rest | 47 | 99.61 |
| **Patro et al. [16]** | ECG-ID | Rest2Rest | 90 | **99.14** |
| Zhao et al. [11] | MIT-BIH | Rest2Rest | 47 | 98.36 |
| Zhao et al. [11] | exercise-ECGID | Rest2Rest | 45 | 99.86 |
| Zhao et al. [11] | ECG-ID | Rest2Rest | 90 | 97.13 |
| **Proposed** | exercise-ECGID | Rest2Rest | 45 | **99.94** |
| **Proposed** | MIT-BIH | Rest2Rest | 47 | **99.79** |
| Proposed | ECG-ID | Rest2Rest | 90 | 97.80 |

### 4.3 Ablation Experiments

This section presents detailed ablation experiments to validate the effectiveness of key components in the proposed method. It systematically analyzes the impact of the multi-scale feature extraction module, deep feature extraction module, and self-attention mechanism on model performance. All ablation experiments were conducted under the Rest2Exercise scenario of the exercise-ECGID dataset. Specifically, starting from the baseline model (A1), key components were progressively removed or replaced to construct five groups of controlled experiments.

Table 4 and Fig. 2 compare the performance of the various model variants in the cross-physiological-state ECG authentication task. Fig. 3. display the t-SNE visualization of feature distributions under full model configurations. Each point represents an ECG segment, and different colors indicate different subjects.

**Table 4.** Ablation Experiment Configuration and Performance Comparison

| Model | Multi-scale | Deep-Conv | Self-attention | Acc(%) | AUC-ROC(%) |
|---|---|---|---|---|---|
| **A1(Full Model)** | √ | √ | √ | **92.50** | **98.72** |
| A2 | × | √ | √ | 90.35 | 97.36 |
| A3 | √ | × | √ | 57.28 | 70.73 |
| A4 | √ | √ | × | 90.74 | 97.60 |
| A5 | × | √ | × | 90.74 | 98.03 |

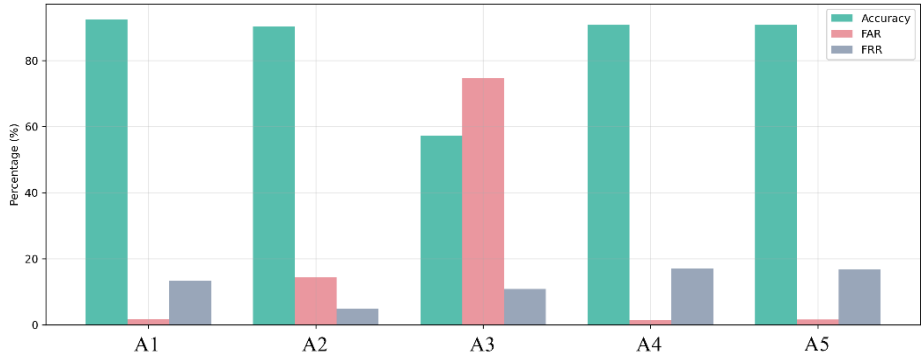

**Fig. 2.** Performance Comparison of Different Model Configurations

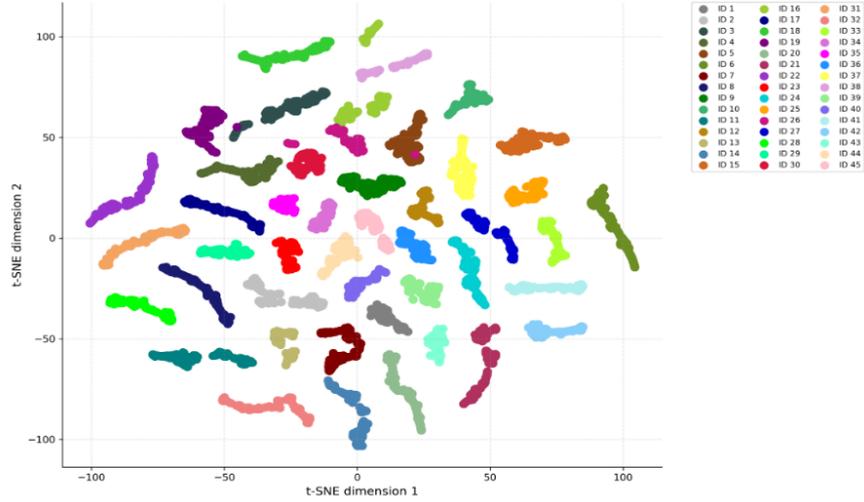

**Fig. 3.** t-SNE Visualization of Feature Distributions under A1 Model Configurations

As shown in Figure 3, the complete model (A1) exhibits the best feature distribution, with tightly clustered samples from the same subject and distinct sample separation from different subjects.

As shown in Table 4, the complete model (A1) achieved the best performance, with an accuracy of 92.5%, AUC of 98.72%, while maintaining a low False Acceptance Rate (FAR) of 1.67% and an acceptable False Rejection Rate (FRR) of 13.33%. Removing the deep convolution module (A3) caused a significant performance drop, with a 35.22% decrease in accuracy and a 28.01% drop in AUC. The importance of the multi-scale feature extraction module is reflected in the 2.15% performance degradation when removed (A2). Similarly, although removing the self-attention mechanism (A4) caused only a slight decrease in performance, it significantly affected the balance between FAR and FRR. Interestingly, the minimal configuration that retained only deep convolution (A5) still maintained robust performance, with a 90.74% balanced accuracy, further confirming the foundational importance of deep convolution in the proposed architecture. These results collectively validate the rationality of the design and demonstrate the complementary effects of the different components in achieving robust ECG authentication under post-exercise and rest conditions.

## 5   Conclusion

This paper proposes CrossStateECG, a multi-scale deep convolution model for ECG biometric recognition across different physiological states. This model addresses the fundamental challenge of cross-state ECG authentication, which has received less attention in existing research. The proposed method effectively extracts discriminative features from ECG signals in rest and post-exercise states by incorporating a deep convolutional network as the core component.

In within-state experiments, the method achieved 99.94% accuracy in the Rest2Rest scenario and 99.86% in the Exercise2Exercise scenario. In mixed-state training (Mix2Mix), the accuracy reached 97.85%. In more challenging cross-state scenarios, the method achieved 92.50% accuracy in the Rest2Exercise and 94.72% in the Exercise2Rest, significantly outperforming existing approaches. Ablation experiments further confirmed the critical role of the deep convolution module, with its removal resulting in a 35.22% decrease in accuracy and a 28.01% drop in AUC. Introducing multi-scale feature extraction and the self-attention mechanism enhanced the model's robustness. Validation experiments on the ECG-ID and MIT-BIH public datasets achieved accuracies of 97.80% and 99.79%, respectively, demonstrating the method's generalization ability. These cross-scenario and cross-dataset results consistently highlight the superiority of the proposed method and provide a practical solution for ECG authentication technology in dynamic real-world environments.

## References

1. Biel, L., Pettersson, O., Philipson, L., Wide, P.: ECG analysis: a new approach in human identification. IEEE Trans. Instrum. Meas. 50, 808–812 (2001). https://doi.org/10.1109/19.930458.
2. Shen, T.W., Tompkins, W.J., Hu, Y.H.: One-lead ECG for identity verification. In: Proceedings of the Second Joint 24th Annual Conference and the Annual Fall Meeting of the Biomedical Engineering Society] [Engineering in Medicine and Biology. pp. 62–63. IEEE, Houston, TX, USA (2002). https://doi.org/10.1109/IEMBS.2002.1134388.
3. Israel, S.A., Irvine, J.M., Cheng, A., Wiederhold, M.D., Wiederhold, B.K.: ECG to identify individuals. Pattern Recognit. 38, 133–142 (2005). https://doi.org/10.1016/j.patcog.2004.05.014.
4. Hejazi, M., Al-Haddad, S.A.R., Singh, Y.P., Hashim, S.J., Abdul Aziz, A.F.: ECG biometric authentication based on non-fiducial approach using kernel methods. Digital Signal Processing. 52, 72–86 (2016). https://doi.org/10.1016/j.dsp.2016.02.008.
5. Liu, F., Si, Y., Luo, T.: The ECG identification based on GRNN.
6. Liu, X., Si, Y., Wang, D.: LSTM Neural Network for Beat Classification in ECG Identity Recognition. AUTOSOFT. 1--1 (2019). https://doi.org/10.31209/2019.100000104.
7. Hsu, P.-Y., Hsu, P.-H., Liu, H.-L.: Fold electrocardiogram into a fingerprint. In: 2020 IEEE/CVF Conference on Computer Vision and Pattern Recognition Workshops (CVPRW). pp. 3612–3619. IEEE, Seattle, WA, USA (2020). https://doi.org/10.1109/CVPRW50498.2020.00422.
8. Liu, X., Si, Y., Yang, W.: A Novel Two-Level Fusion Feature for Mixed ECG Identity Recognition. Electronics. 10, 2052 (2021). https://doi.org/10.3390/electronics10172052.
9. Yang, Z., Liu, L., Li, N., Tian, J.: ECG identity recognition based on feature reuse residual network. Processes. 10, 676 (2022). https://doi.org/10.3390/pr10040676.
10. Zhang, X., Liu, Q., He, D., Suo, H., Zhao, C.: Electrocardiogram-Based Biometric Identification Using Mixed Feature Extraction and Sparse Representation. Sensors. 23, 9179 (2023). https://doi.org/10.3390/s23229179.
11. Zhao, Y., Zhao, L., Xiao, Z., Li, J., Liu, C.: Enhancing Electrocardiogram Identity Recognition Using Convolutional Neural Networks With a Multisimilarity Loss Model. IEEE Trans. Instrum. Meas. 73, 1–8 (2024). https://doi.org/10.1109/TIM.2024.3368481.

12. Wang, X., Han, X., Huang, W., Dong, D., Scott, M.R.: Multi-similarity loss with general pair weighting for deep metric learning. In: 2019 IEEE/CVF Conference on Computer Vision and Pattern Recognition (CVPR). pp. 5017–5025. IEEE, Long Beach, CA, USA (2019). https://doi.org/10.1109/CVPR.2019.00516.
13. Moody, G. B., Mark, R. G.: The impact of the MIT-BIH Arrhythmia Database. IEEE Eng in Med and Biol 20(3), 45-50 (May-June 2001). (PMID: 11446209)
14. Wang, Z., Li, Y., Cui, W.: ECG identification under exercise and rest situations via various learning methods, http://arxiv.org/abs/1905.04442, (2019).
15. Kim, B.-H., Pyun, J.-Y.: ECG identification for personal authentication using LSTM-based deep recurrent neural networks. Sensors. 20, 3069 (2020). https://doi.org/10.3390/s20113069.
16. Patro, K.K., Reddi, S.P.R., Khalelulla, S.K.E., Rajesh Kumar, P., Shankar, K.: ECG data optimization for biometric human recognition using statistical distributed machine learning algorithm. J Supercomput. 76, 858–875 (2020). https://doi.org/10.1007/s11227-019-03022-1.
17. Sung, D., Kim, J., Koh, M., Park, K.: ECG authentication in post-exercise situation. In: 2017 39th Annual International Conference of the IEEE Engineering in Medicine and Biology Society (EMBC). pp. 446–449. IEEE, Seogwipo (2017). https://doi.org/10.1109/EMBC.2017.8036858.
18. Lugovaya, T. S.: Biometric human identification based on electrocardiogram. [Master's thesis] Faculty of Computing Technologies and Informatics, Electrotechnical University "LETI", Saint-Petersburg, Russian Federation; June 2005.
19. Goldberger, A., Amaral, L., Glass, L., Hausdorff, J., Ivanov, P. C., Mark, R., Mietus, J. E., Moody, G. B., Peng, C. K., Stanley, H. E.: PhysioBank, PhysioToolkit, and PhysioNet: Components of a new research resource for complex physiologic signals. Circulation 101, e215–e220 (2000).
20. Lin, T.-Y., Goyal, P., Girshick, R., He, K., Dollar, P.: Focal loss for dense object detection. IEEE Trans. Pattern Anal. Mach. Intell. 42, 318–327 (2020). https://doi.org/10.1109/TPAMI.2018.2858826.
21. Saleh, A.A., Sprecher, E., Levy, K.Y., Lange, D.H.: DE-PADA: Personalized augmentation and domain adaptation for ECG biometrics across physiological states, http://arxiv.org/abs/2502.04973, (2025). https://doi.org/10.48550/arXiv.2502.04973.
22. Kim, H., Phan, T.Q., Hong, W., Chun, S.Y.: Physiology-based augmented deep neural network frameworks for ECG biometrics with short ECG pulses considering varying heart rates. Pattern Recognition Letters. 156, 1–6 (2022). https://doi.org/10.1016/j.patrec.2022.02.014.
23. Fuster-Barceló, C., Cámara, C., Peris-López, P.: ECG-based patient identification: A comprehensive evaluation across health and activity conditions, http://arxiv.org/abs/2302.06529, (2024). https://doi.org/10.48550/arXiv.2302.06529.
24. Zhang, X., Liu, Q., He, D., Suo, H., Zhao, C.: Electrocardiogram-Based Biometric Identification Using Mixed Feature Extraction and Sparse Representation. Sensors. 23, 9179 (2023). https://doi.org/10.3390/s23229179.